\documentclass{article}

    \usepackage[preprint]{neurips_2024}

\usepackage[utf8]{inputenc} 
\usepackage[T1]{fontenc}    
\usepackage{url}            
\usepackage{booktabs}       
\usepackage{amsfonts}       
\usepackage{nicefrac}       
\usepackage{microtype}      
\usepackage{xcolor}         
\usepackage{hyperref}
\usepackage{graphicx}
\usepackage{subfigure}
\usepackage{amsmath}
\usepackage{amssymb}
\usepackage{mathtools}
\usepackage{amsthm}

\usepackage{amsmath,amsfonts,bm}

\def\eqref#1{equation~\ref{#1}}

\def\1{\bm{1}}

\def\ve{{\bm{e}}}

\def\evp{{p}}

\def\evu{{u}}

\def\evx{{x}}

\DeclareMathAlphabet{\mathsfit}{\encodingdefault}{\sfdefault}{m}{sl}
\SetMathAlphabet{\mathsfit}{bold}{\encodingdefault}{\sfdefault}{bx}{n}

\def\gE{{\mathcal{E}}}

\def\gG{{\mathcal{G}}}

\def\gQ{{\mathcal{Q}}}

\def\gV{{\mathcal{V}}}

\usepackage[capitalize,noabbrev]{cleveref}

\title{Leveraging Temporal Graph Networks Using Module Decoupling}

\author{%
  Or Feldman
  \\
  Department of Computer Science\\
  Technion – Israel Institute of Technology\\
  \texttt{orfeldman@campus.technion.ac.il} \\
   \And
  Chaim Baskin\\
  Department of Computer Science\\
  Technion – Israel Institute of Technology\\
  \texttt{chaimbaskin@technion.ac.il} \\
}

\begin{document}

\maketitle

\begin{abstract}
Current memory-based methods for dynamic graph learning utilize batch processing to handle rapid updates efficiently. However, the adoption of batches introduces a phenomenon we term as \textit{missing updates}, which adversely affects the performance of memory-based models. In this work, we analyze the impact of \textit{missing updates} on dynamic graph learning models and propose a decoupling strategy to mitigate these effects. Implementing this strategy, we present the Lightweight Decoupled Temporal Graph Network, a memory-based model with a minimal number of learnable parameters that is capable of dealing with the demand of high frequency of updates.
We validated our approach across diverse dynamic graph benchmarks. LDTGN surpassed the average precision of previous methods by over 20\% in scenarios demanding frequent graph updates, such as US Legis or UN Trade. In the vast majority of the benchmarks, LDTGN achieves state-of-the-art or comparable results while operating with significantly higher throughput than existing baselines. The code to replicate our experiments is available at \href{https://anonymous.4open.science/r/Modules-Decoupling-TGN-6FE7/}{this url}. 
\end{abstract}
\section{Introduction}

Dynamic graphs are commonly used to describe real-world dynamic systems, where the interacting elements are modeled as nodes, and the interactions between two elements are represented as edges. Each edge is usually labeled with a timestamp indicating its time of occurrence. Item recommendation on e-commerce platforms \citep{ding2019user}, friendship suggestion on social networks \citep{backstrom2011supervised,haghani2019systemic}, anomaly detection on communication networks \citep{yu2018netwalk}, and traffic forecasting \citep{cini2023scalable} are all practical tasks that can be modeled using dynamic graphs.

Although most graph-related real-world tasks are time-evolving, deep learning approaches usually focus on problems described using static graphs, which do not change over time. Moreover, it has also been shown that ignoring the dynamic nature of a system by abstracting it with static graphs is suboptimal \citep{rossi2020temporal,xu2020inductive}. A dynamic representation of a system, on the other hand, is often able to define the evolving behavior of the latter \citep{simmel1950sociology,granovetter1973strength,mangan2003structure,toivonen2007role,gorochowski2018organization}.

Dynamic graph approaches are often based on discrete-time \citep{liben2003link,sankar2020dysat,pareja2020evolvegcn} or continuous-time \citep{trivedi2019dyrep,ma2020streaming,cong2023we} settings. In discrete-time settings, data are received as a sequence of snapshots describing the full graph structure at specific times, while in the flexible continuous-time setting, a single update on the graphs can happen at any moment.
The setting in which deep learning models for dynamic graphs operate at inference time can be roughly divided into the following types: streaming, deployed, and live update \citep{huang2024temporal}. In this work, we focus on continuous-time dynamic graphs in the streaming context, in which the models may be updated upon receiving new information, but cannot perform backpropagation due to the high throughput required.  

Memory-based models for dynamic graphs are designed to support the assimilation of new information through graph updates during the inference phase. To do this, they manage a memory unit that represents the current state of the dynamic graph.
This memory unit usually includes the current structure of the dynamic graph, data-specific information such as node and edge features, timestamps of previous updates, and learnable information computed by the model.

In the streaming setting for continuous-time dynamic graphs, memory-based deep networks have to use batches to keep up with the stream of incoming updates, which means they process multiple updates in parallel. This situation introduces a new problem in which updates for the models are not being considered for the predictions of inputs inside their mutual batch. In \cref{sec:problem_statement}, we formally define this undesirable phenomenon as \textit{missing updates}. In this work, we suggest a decoupling strategy that minimizes the negative impacts of \textit{missing updates} while still using batches. Guided by this strategy, we have built the Lightweight Decoupled Temporal Graph Network (LDTGN) -- an efficient memory-based model for dynamic graph learning that outperforms most established baselines both in terms of running time and performance.

To summarize, this work makes the following contributions: 
\begin{itemize}
    \item We introduce and analyze the problem of \textit{missing updates} when using memory-based models.
    \item We suggest a novel decoupling methodology for building deep learning memory-based models for dynamic graph learning.
    \item Based on the suggested methodology, we propose a new lightweight model for dynamic graph learning tasks that can operate at high streaming rates and with significantly smaller number of parameters compared to other baselines.
    \item We evaluated LDTGN on various transductive and inductive benchmarks for dynamic graph learning and achieved state-of-the-art or comparable performance on most of the tested benchmarks, while outperforming previous methods in terms of throughput.
\end{itemize}
\section{Background}
\label{sec:background}
Static graph $\gG=(\gV,\gE)$ is a tuple of vertex set $\gV$ and edge set $\gE$, s.t., $\ve\in\gE$ is a tuple of two vertices from $\gV$. 
$\gG$ is often equipped with features functions $F_\gV: \gV\rightarrow \mathbb{R}^n$ and $F_\gE: \gE\rightarrow\mathbb{R}^n$ that maps a vertex or an edge into an $n$-dimensional vector representing their matching features.
Continuous-Time Dynamic Graph (CTDG) is a sequence $\gQ = \{\evu_{t_1},\evu_{t_2},...,\evu_{t_m}\}$ of $m$ timestamped updates on the graph. An update $\evu_t$ that occurs at time $t$ can be one of the following: node addition, node removal, edge addition, and edge removal. $\gG[\gQ(t)]$ is the static graph received by applying all the updates from $\gQ$ on $\gG$ that have occurred until time $t$.
The $k$-hop neighborhood of a node $v_i$ at time $t$ is defined by:
\begin{gather}
\mathcal{N}_i^0(t) = \{v_i\} \\
\mathcal{N}_i^k(t) = \{v_j | v_u \in \mathcal{N}_i^{k-1}(t) , (v_u,v_j) \in \gG[\gQ(t)]\}
\end{gather}

As a result of the growing interest in CTDG  with a stream of updates, several techniques were recently developed  \citep{kumar2019predicting,trivedi2019dyrep,xu2020inductive,wang2021tcl,wang2021inductive,cong2023we}. Many of these methods are specific cases of the Temporal Graph Network \citep[TGN,][]{rossi2020temporal} model. TGN is a general memory-based deep learning architecture designed to learn on CTDG while achieving throughput suitable for streaming tasks. The primary concept of TGN is to maintain states, namely node features, which are updated with each modification to the graph. To achieve this objective, TGN utilizes two central modules: memory and prediction.

\paragraph{Memory module} The memory module is responsible for applying the updates on the graph and update the states accordingly. When a new batch of updates arrives, the memory module applies a message function that generates a vector for each node involved in each update. If the update is the addition of edge $e_{i,j}$ from node $v_i$ to node $v_j$ at time $t$, then the appropriate messages of $v_i$ and $v_j$ are:
\begin{gather}
 m_i(t)=\mathrm{msg_s}(s_i(t^-),s_j(t^-),\Delta t_i,F_{\gE}(e_{i,j})) \\ m_j(t)=\mathrm{msg_d}(s_j(t^-),s_i(t^-),\Delta t_j,F_{\gE}(e_{i,j}))  
\end{gather}

where $s_u(t^-)$ is the state of $v_u$ prior to $t$ and, $\Delta t_u$ is  the time elapsed since $v_u$ received an update. $\mathrm{msg_s}$ and $\mathrm{msg_d}$ may have learnable parameters.
Then, all the messages in the batch are aggregated into a single message per node, s.t., if node $v_i$ is involved in updates at times $t_1\leq t_2 \leq...t_n=t$ then:
\begin{equation}
    \overline{m}_i(t)= \mathrm{agg}(m_i(t_1),m_i(t_2)...m_i(t_n))
\end{equation}
The aggregation function, for example, can take only $m_i(t_n)$ and neglects any previous messages in the batch. 
Finally, the message updater updates the state of the nodes:
\begin{equation}
    s_i(t)=\mathrm{mem}(\overline{m}_i(t),s_i(t^-))
\end{equation}
The $\mathrm{mem}$ function is a memory-based neural network such as LSTM \citep{hochreiter1997long} or GRU \citep{cho2014learning}.  

\paragraph{Prediction module}
The prediction module computes the predictions for the inputs in a given batch, e.g., potential edges in a link prediction task. First, the prediction module reads from the memory module all the states of the nodes in the neighborhood of any node involved in the input. Then it generates a new embedding for each node based on its state and the states of its neighbors. For example, denote $[\boldsymbol{\cdot}||\boldsymbol{\cdot}]$ as the operation of vector concatenation, then the embedding formulation based on $1$-hop neighborhood of node $v_i$ is:
\begin{equation}
    z_i(t) = \Sigma_{v_j\in\mathcal{N}_i^1(t)} h(v_i(t),v_j(t),F_{\gE}(e_{i,j}))
\end{equation}
where $v_u(t) = [F_\gV(v_u)||s_u(t^-)||\Delta t_u]$  and $h$ is a learnable function. Using the neighborhood of a node in the graph to compute its embedding averts the staleness problem\citep{kazemi2020representation}.

For the task of future link prediction of edge $e_{i,j}$ at time $t$, TGN computes the edge's probability to exist by:
\begin{equation}
    p_{i,j}(t)=\mathrm{merge}(z_i(t),z_j(t))
\end{equation}
where $\mathrm{merge}$ is a learnable function such as an $\mathrm{MLP}$.

\section{Problem statement}
\label{sec:problem_statement}
The processing sequence of memory-based models for dynamic graph learning involves receiving a new batch containing both graph updates and inputs for prediction. This utilization of batches allows memory-based models to achieve a reasonable throughput during inference time \citep{kumar2019predicting,rossi2020temporal,wang2021inductive,cong2023we}. In the streaming setting, where the graph receives new updates at extremely
high speeds, it is crucial for the model to have sufficient
throughput. Otherwise, a buffer to the model containing the new updates will overflow. 

Memory-based models have a well-defined flow of operation upon receiving a new batch. Initially, they compute their predictions for the inputs in the batch. This operation is performed in parallel by using the current states and the current graph structure as saved in their memory. Subsequently, they process all the updates in the batch and update their inner memory accordingly. This flow of operations introduces the undesirable phenomenon defined as \textit{missing updates}.

Formally, given a batch $\gQ=\{\evx_{t_1},\evx_{t_2}...\evx_{t_m}\}$ of size $m$ where $\evx_{t_i}$ can be an update to the graph $\evu_{t_i}$ or an input to predict $\evp_{t_i}$.  We say that $\evu_{t_i}$ is a \textit{missing update} if there exists an input $\evp_{t_j}$, s.t., $i<j$ and the nodes involved in $\evu_{t_i}$ are in the neighborhood of the nodes involved in $\evp_{t_j}$. Inputs to the model that depend on \textit{missing updates} are harder to predict since the memory of the model of their neighborhood is outdated at the time of the prediction. 

\subsection{Empirical analysis}
\begin{figure*}
\centering     
\subfigure[]{\label{fig:a}\includegraphics[width=46mm]{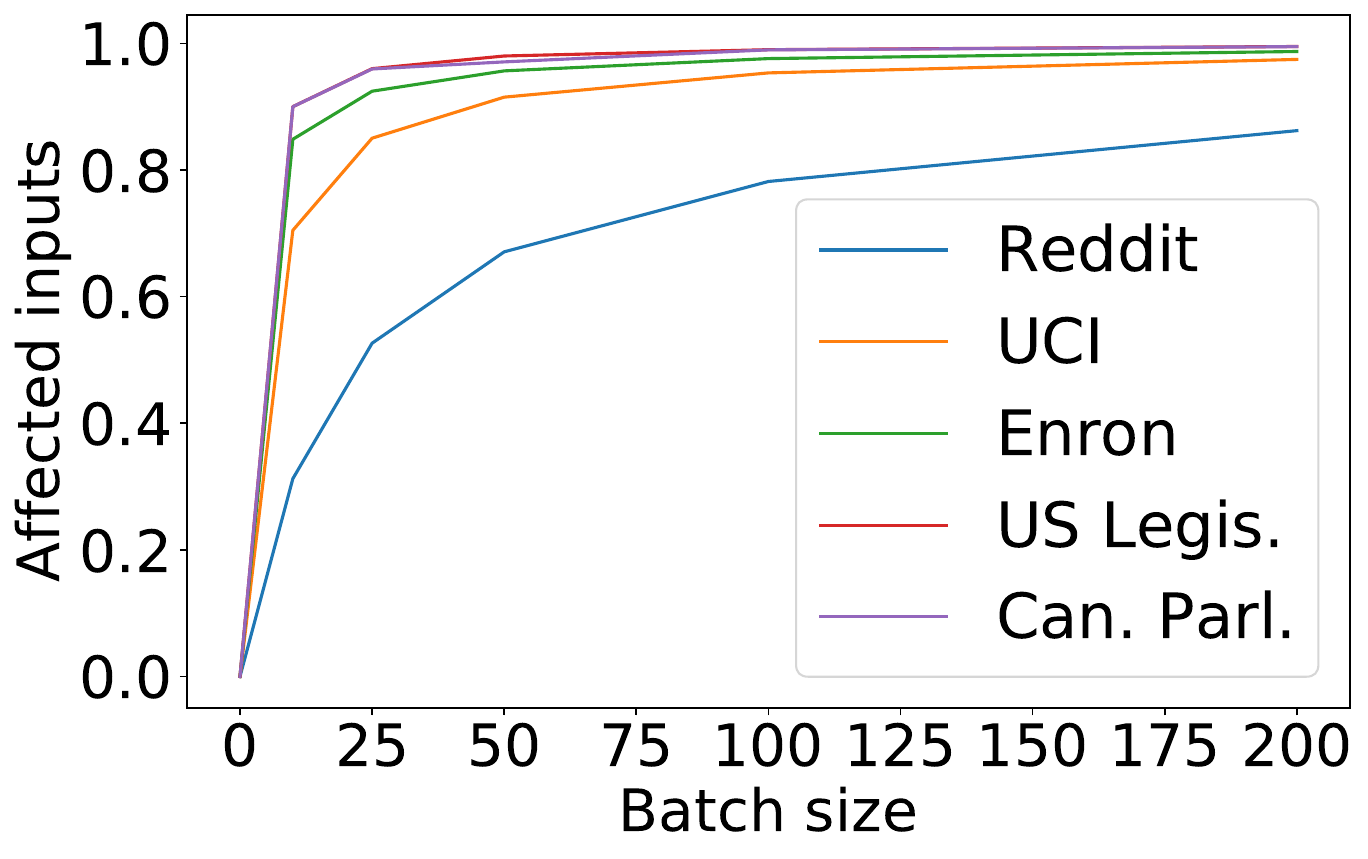}}
\subfigure[]{\label{fig:b}\includegraphics[width=46mm]{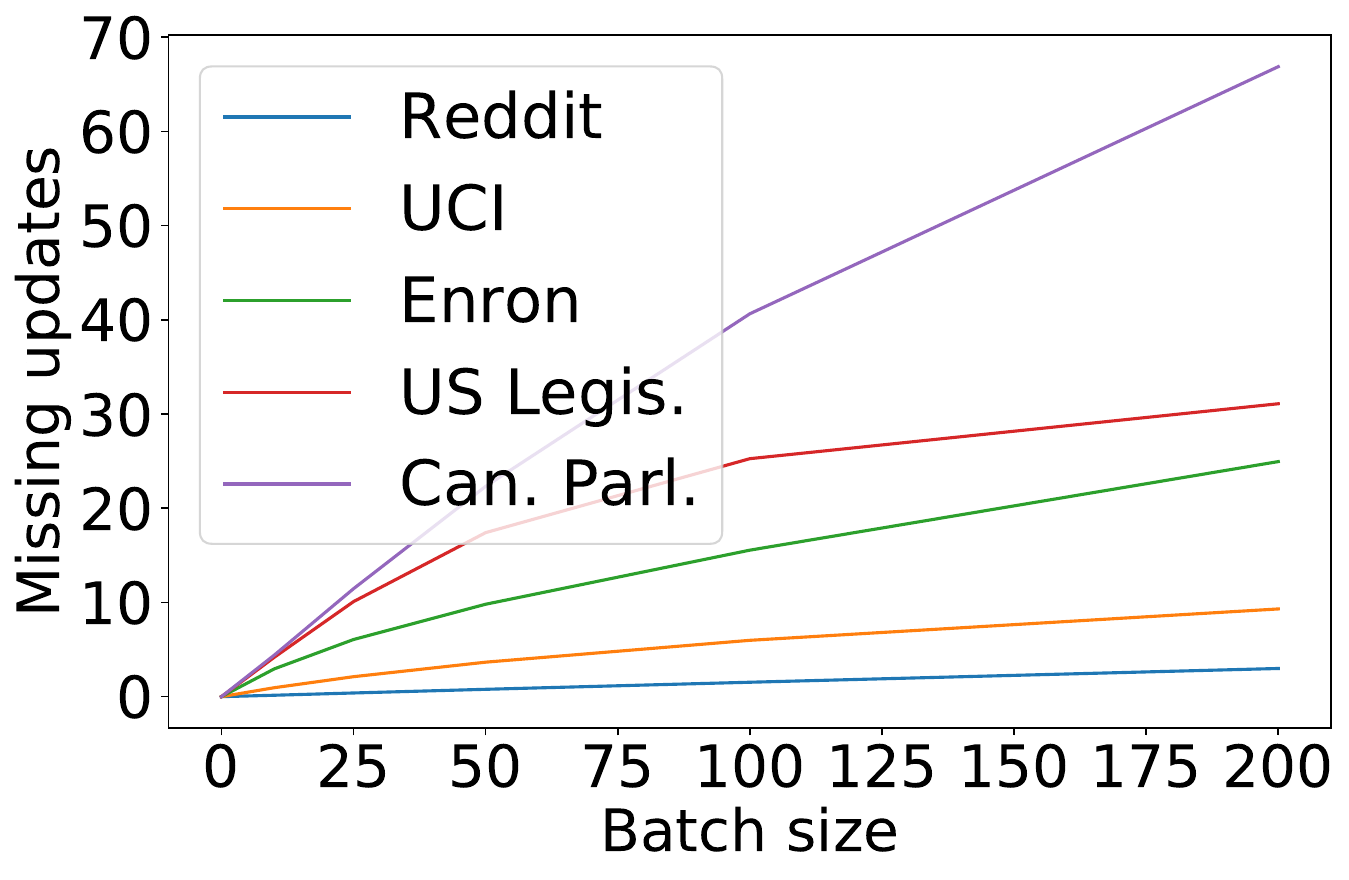}}
\subfigure[]{\label{fig:c}\includegraphics[width=46mm]{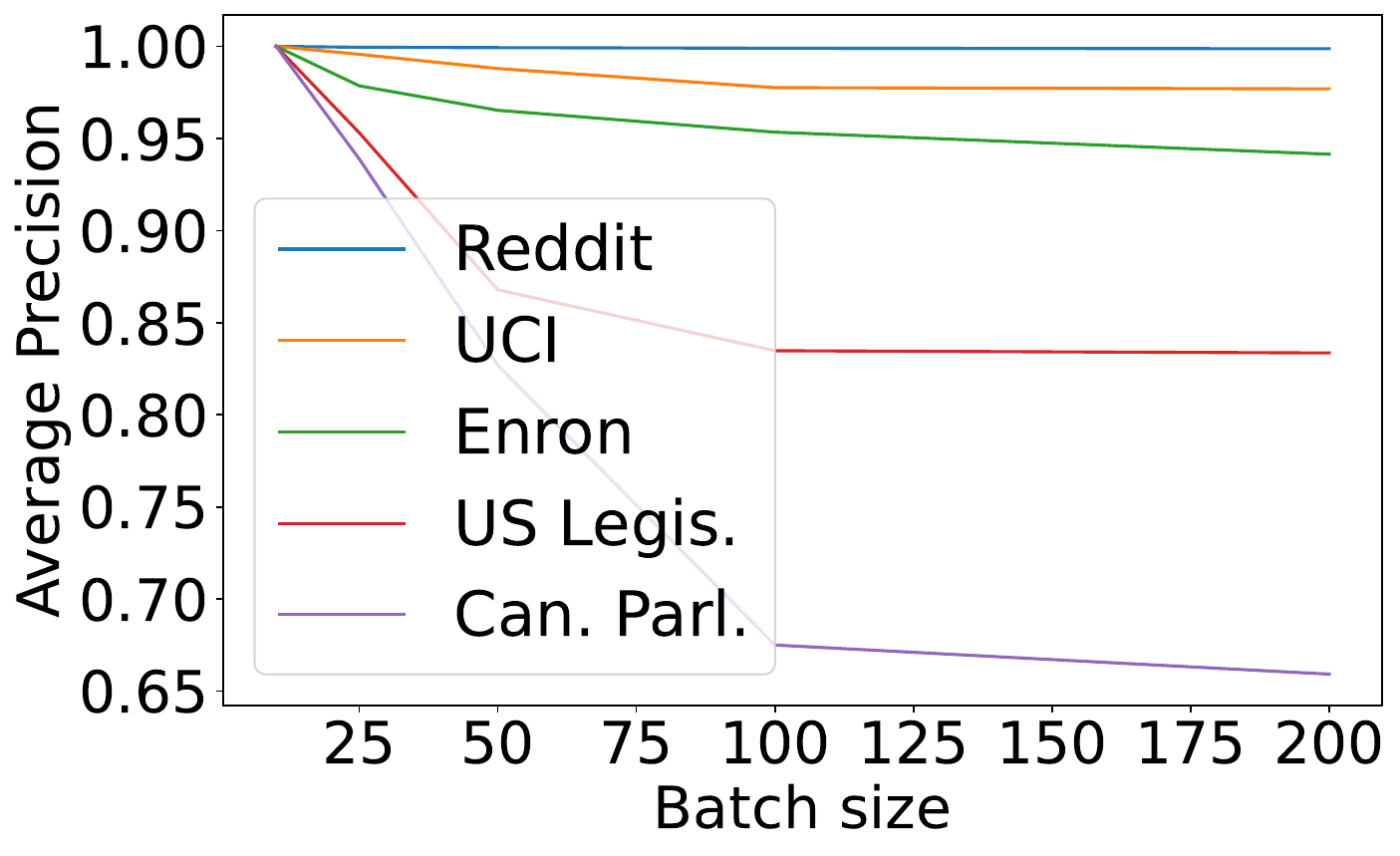}}
\caption{ The incidence of \textit{missing updates} in real-world datasets as a function of the batch size and their impact on the performance of TGN. In \subref{fig:a}, the ratio of inputs that depend on at least a single \textit{missing update} increases significantly as the batch size increases. In \subref{fig:b}, the average number of \textit{missing updates} per input increases as the batch size increases. In \subref{fig:c}, the performance of TGN corresponds to the extent of \textit{missing updates}, where a high incidence of \textit{missing updates} indicates a significant performance decrease.}
\label{fig:missing_updates}
\end{figure*}

We examined the incidence and impact of the \textit{missing updates} phenomenon on various real-world datasets for dynamic graphs. We measured the average ratio of inputs in a batch that depend on at least a single \textit{missing update}. In addition, we tested the average number of \textit{missing updates} that affects a single input to the model. In both cases, we examined \textit{missing updates} within the $1$-hop neighborhood of the input nodes for different batch sizes and reported the results in \cref{fig:a} and \cref{fig:b} respectively. We also tested a standard TGN trained on these datasets with different batch sizes. In \cref{fig:c}, we report the average precision of TGN for each batch size, normalized by the average precision achieved with a batch size of 10. In \cref{appendix:A}, we supply the full \textit{missing updates} statistics for all the datasets used in this work.

In \cref{fig:missing_updates}, it is observed that an increase in batch size correlates with a higher incidence of \textit{missing updates}. Furthermore, the occurrence of \textit{missing updates} varies across different datasets. The findings from \cref{fig:missing_updates} suggest a negative connection between the occurrence of \textit{missing updates} in a dataset and the performance of a model trained on it. Consequently, achieving optimal performance in memory-based models necessitates a smaller batch size. This result indicates a trade-off with respect to the batch size in memory-based models, as large batch size is required to attain high throughput in the streaming scenario.

\section{Related work}

\paragraph{Handling \textit{missing updates}} 
The t-Batch algorithm \citep{kumar2019predicting} was initially intended to improve the running-time performance of memory-based deep networks for dynamic graphs that process updates one after
the other (i.e., batch size equal to 1). The logic motivating t-Batch is that these networks can combine multiple updates into a single batch and apply them in parallel if they do not contain the same nodes, where the batches are temporally sorted. Using t-Batch, JODIE's memory-based model becomes X9.2 faster than similar methods without suffering from \textit{missing updates} \citep{kumar2019predicting}.
The t-batch algorithm, however, suffers from two main flaws. First, large batch sizes for t-Batch are often impossible since temporal locality is a common characteristic of dynamic graphs \citep{poursafaei2022towards}. In addition, many modern deep learning networks for dynamic graphs, such as TGN, depend on the neighborhood of the nodes to give an appropriate prediction, causing t-Batch to perform complicated neighborhood-independent batches instead of node-independent batches, which are significantly smaller. 
\paragraph{Efficient methods for streaming} According to \citet{huang2024temporal}, EdgeBank is currently an order of magnitude faster than other well-known techniques for dynamic graphs. EdgeBank \citep{poursafaei2022towards} is a memorization algorithm that saves any seen update and predicts according to a simple decision rule that can be one of the following: whether the input was seen in the last few iterations or whether the input has already been seen a sufficient number of times.
The algorithm's simplicity allows it to perform extremely fast, even without batches, thus not suffering from \textit{missing updates}. Nevertheless, EdgeBank was developed to serve as a baseline for testing and comparing other methods for dynamic graphs \citep{poursafaei2022towards}, and, therefore, its performance lags significantly behind the state-of-the-art \citep{yu2023towards,huang2024temporal}.

\section{Proposed method}
\label{sec:proposed_method}
In this section, we describe our method to balance the batch size trade-off discussed earlier in \cref{sec:problem_statement}. The method decouples the TGN modules, by ensuring that each module uses a different batch size. In general, the memory module will utilize smaller batch sizes for frequent updates, while the prediction module will employ larger batch sizes for efficiency.

Following that, we describe our proposed lightweight model for dynamic graph learning tasks. The model is a TGN with decoupled modules implemented using efficient functions. Specifically, we parameterize the EdgeBank \citep{poursafaei2022towards} model to allow it to learn. Then, we add extra parameters to consider single-node information in the prediction instead of solely relying on edge temporal information.   

\subsection{The decoupling strategy}
\label{subsection:decoupling}
We propose to \textit{decouple} the core modules of TGN: the prediction and memory modules. The decoupled modules will operate on different data and different batch sizes. Given a batch containing updates to apply and inputs to predict, the model divides the batch into smaller consecutive batches termed memory batches. The memory module operates on the memory batches, and thus, it can perform memory updates more frequently. After processing a memory batch but before proceeding to the next one, the memory module extracts and temporally saves the temporal neighborhood information. This information encompasses the neighborhood state relevant to the nodes in the subsequent memory batch, preventing it from being overridden. The neighborhood state is defined by:
\begin{equation}
    S_{\mathcal{N}_i^k(t)}(t) = \{s_j(t)\ | v_j\in \mathcal{N}_i^k(t)\}
\end{equation}

This creates a view of the model's memory for each time a memory batch starts. After processing all memory batches, the prediction module reads the extracted states for each node associated with a given input from the view before that input's timestamp. Subsequently, the prediction module simultaneously computes predictions for all inputs within the complete batch.

\begin{figure}[!htb]
   \begin{minipage}{0.48\textwidth}
     \centering
     \includegraphics[width=.98\linewidth]{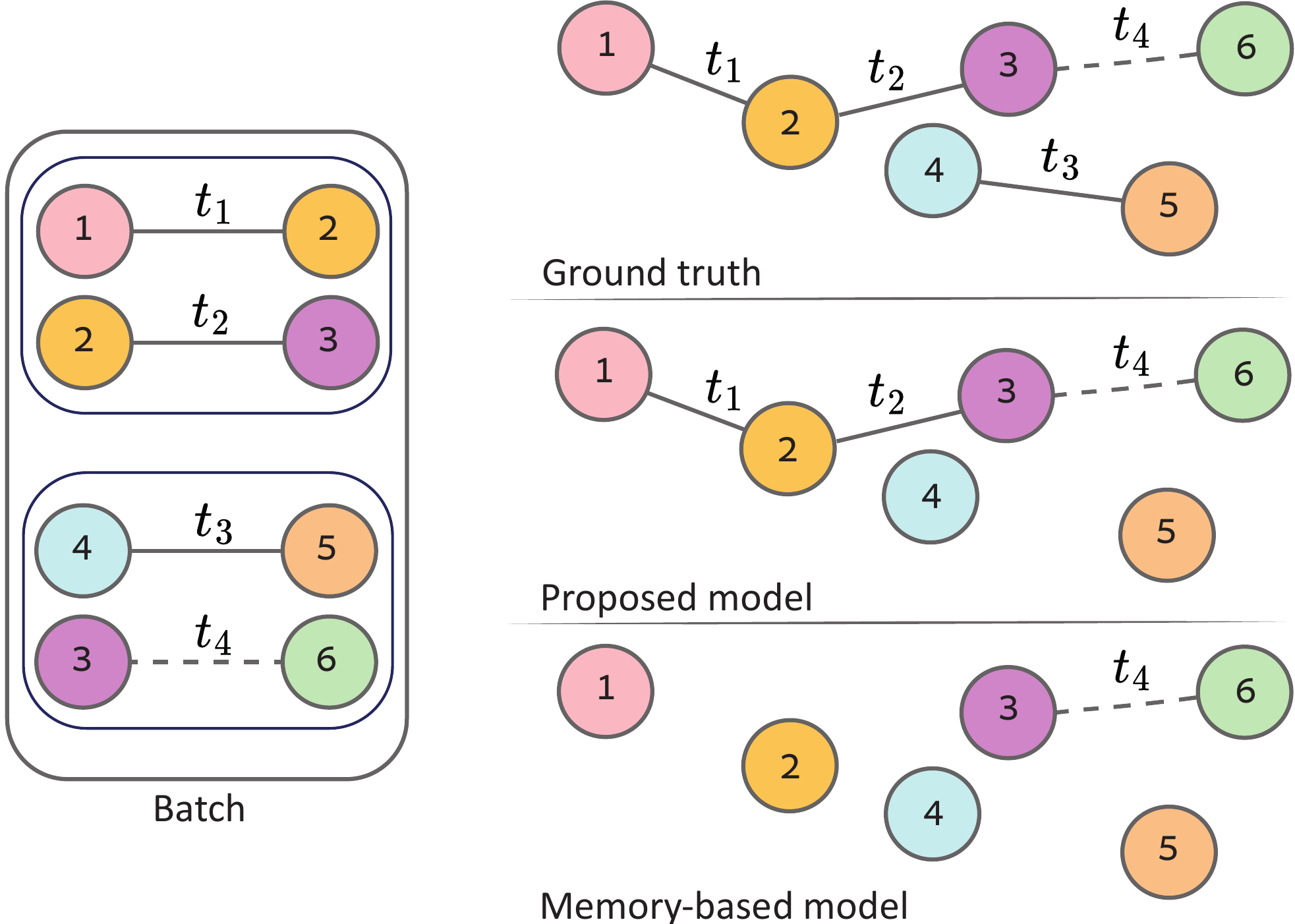}
     \caption{Illustration of a dynamic graph at $t_4$ for the task of predicting the edge $(v_3,v_6)$. The state of a memory-based model is compared to the state of a model operating using the proposed decoupling strategy. The memory-based model was updated prior to $t_{1}$ and, therefore, does not contain $(v_1,v_2)$,$(v_2,v_3)$ and $(v_4,v_5)$.
The model that follows the decoupling strategy and applies inner batch updates was previously updated at $t_{2}$ and, therefore, closely resembles the ground truth and is missing only $(v_4,v_5)$.}
\label{fig:teaser}
   \end{minipage}\hfill
   \begin{minipage}{0.48\textwidth}
     \centering
     \includegraphics[width=.98\linewidth]{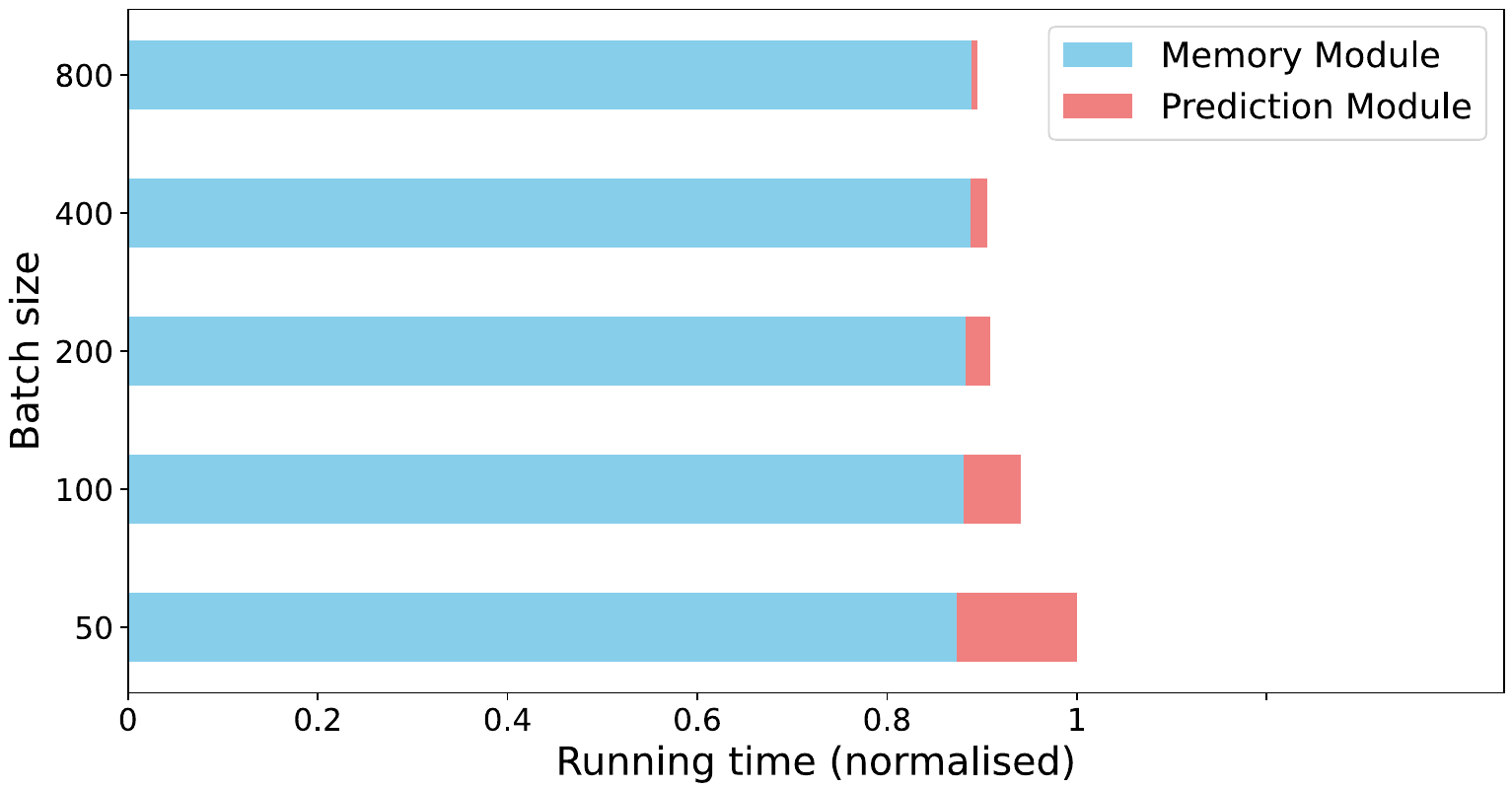}
     \caption{Comparison of running times for decoupled TGN with a constant memory batch size of 50 and varying batch sizes on the test set of the Wikipedia dataset. The running times are normalized by the baseline scenario where both the memory batch size and the batch size are set to 50.} 
     \label{fig:decoupled_running_time}
   \end{minipage}
\end{figure}

\Cref{fig:teaser} demonstrates the effectiveness of a decoupled model compared to a standard memory-based model. The edge at $t_4$ in \Cref{fig:teaser} is given as an input for the models. A standard memory-based model computes the embeddings of $v_3$ and $v_6$ based on their neighborhood states before $t_1$ and only then updates its inner memory with the edges at $t_1$,$t_2$ and $t_3$. On the other hand, a decoupled model initially performs memory updates of the two memory batches. Then, the prediction module uses the states extracted before $t_3$ that include the updates in first memory batch. In \cref{fig:teaser}, the \textit{missing update} that affects the interaction between $v_3$ and $v_6$ is avoided by using the decoupling strategy since the prediction module is aware of the interaction between $v_2$ and $v_3$.

Decoupling the modules of TGNs offers two immediate benefits. First, by decoupling the memory module from the prediction module and setting the memory batch size to 1, we completely solve the \textit{missing updates} problem. Secondly, we can accelerate the execution time of an existing model without compromising its accuracy. This can be achieved by decoupling its modules, setting the memory batch size to match the model's original batch size, and substantially increasing the new batch size of the prediction module. Using the original batch size for the memory batches ensures the same frequency of \textit{missing updates}, and the new larger batch size will improve the running time performance. \cref{fig:decoupled_running_time} details the running time improvement of decoupled TGN with a memory batch size of 50 when using growing batch sizes. Notably, the decoupling strategy enhances TGN's running time by 12.5\% without compromising its performance, as the frequency of \textit{missing updates} depends only on the memory batch size. Furthermore, actively transferring additional computations from the memory module to the prediction module will lead to an additional improvement in running time. Further analysis of the potential speedup of the decoupling strategy is discussed in \cref{appendix:E}.

\subsection{Lightweight Decoupled Temporal Graph Network}
\label{subsec:ldtgn}
\begin{figure*}[ht]
\vskip 0.2in
\begin{center}
\centerline{\includegraphics[width=0.97\linewidth]{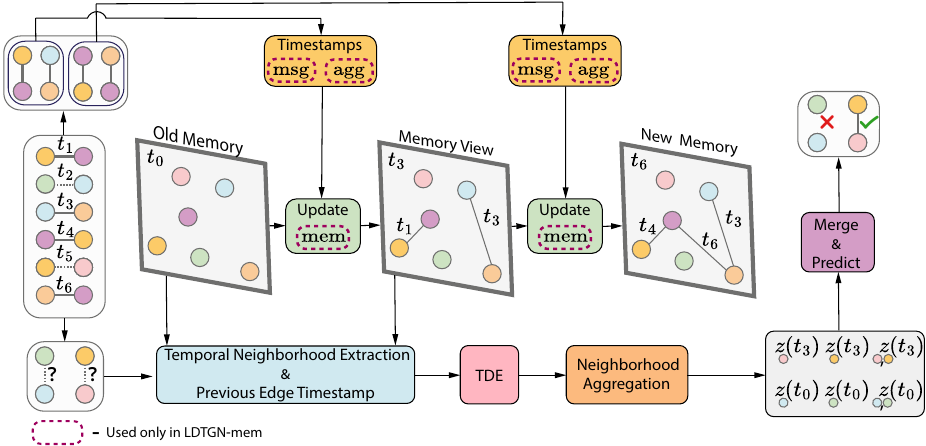}}
\caption{Framework of the proposed model. The batch of updates and inputs is first divided into memory batches and a single batch of inputs. Then, the new edges and their appropriate timestamps are saved in the memory. In LDTGN-mem, the state of each node in the memory batch is updated using the $\mathrm{msg}$, $\mathrm{agg}$, and $\mathrm{mem}$ functions. Before each update, the relevant information is saved in a memory view to prevent it from being overridden. Next, the information of each input node is extracted from the appropriate memory view. Then, $\mathrm{TDE}$ is applied to the time differences between the inputs and the time of the extracted timestamps. Neighborhood information is aggregated using learnable attention weights to create a single encoding for each node. Finally, the nodes encoding and the edge encoding are merged using the $\mathrm{merge}$ function, and the combined encoding is used to get the final prediction.}
\label{fig:framework}
\end{center}
\vskip -0.2in
\end{figure*}
We propose the Lightweight Decoupled Temporal Graph Network (LDTGN), an efficient model designed for dynamic graph learning tasks. LDTGN operates with high throughput, crucial for the streaming setting, while also achieving superior performance in dynamic graph learning tasks.

In this subsection, we develop LDTGN step-by-step by enhancing EdgeBank and incorporating the decoupled strategy. Despite EdgeBank's performance falling short of the current state-of-the-art, it demonstrates commendable results with exceptionally high throughput, making it a suitable foundation for our model. We proceed to delineate the deficiencies in EdgeBank that need addressing to attain top-tier performance. Subsequently, we integrate improvements with EdgeBank to resolve these issues effectively, thereby constructing the LDTGN model. For the simplicity of the presentation, we describe LDTGN for future edge prediction tasks and assume only edge addition updates. Comprehensive details about applying node addition, node removal, and edge removal updates, as well as adjustments for node classification task, are provided in \cref{appendix:C}.
 
The EdgeBank model can be formulated as a memory-based algorithm as presented by \citet{poursafaei2022towards} where an edge that did not get an update in the past $T$ updates is considered negative. We can also describe this memory-based prediction rule as a linear function that maps a time-based difference into a prediction. \cref{eq:edgebank} details the linear function of EdgeBank with a decision function that considers any edge $e_{i,j}$ that was updated in the last $T$ updates as positive. 
\begin{equation}
\label{eq:edgebank}
    p_{i,j}(t) = -(t-t_{i,j}) + T
\end{equation}

In \cref{eq:edgebank}, $t_{i,j}$ is the last time the edge $e_{i,j}$ received an update, and $t$ is the current time. $t_{i,j}$ is set to $0$ if $e_{i,j}$ has not been received yet. \citet{poursafaei2022towards} suggested to use a constant value of 1000 for $T$. The equation should be parameterized to allow the model to learn the most appropriate value of $T$ for every dataset. To do this, we added 
 a bias $b$ and a coefficient $w$ as detailed in \cref{eq:edgebank_param}.
\begin{equation}
\label{eq:edgebank_param}
    p_{i,j}(t) = (t-t_{i,j})w + b
\end{equation}
Using \cref{eq:edgebank_param}, we can learn the suitable threshold for each dataset. As in EdgeBank, this function does not incorporate the nodes themselves into the prediction.
This can easily be solved by adding the time differences of each node in the potential edge and appropriate coefficients as in \cref{eq:linear_nodes}.
\begin{equation}
    \label{eq:linear_nodes}
    p_{i,j}(t) = (t-t_{i,j})w_1 + (t-t_i)w_2 + (t-t_j)w_3 + b
\end{equation}
In \cref{eq:linear_nodes}, $t_i$ and $t_j$ are the last times the nodes $v_i$ and $v_j$ received an update, respectively. 
\cref{eq:linear_nodes} is missing topological and data-specific information such as node and edge features. Moreover, the prediction function is linear, which often causes the learned function to be distant from the ground truth prediction function. To solve this issue, we first create embeddings for the nodes in the potential edge and a preliminary embedding for the edge itself as in \cref{eq:node_embedding,eq:edge_embedding}:
\begin{equation}
\label{eq:node_embedding}
    z_i(t) = \Sigma_{k\in\mathcal{N}_i^1(t)} \alpha_k [ v_i(t)||v_k(t)||F_{\gE}(e_{i,k})]
\end{equation}
\begin{equation}
\label{eq:edge_embedding}
    z_{i,j}(t) = \mathrm{TDE}(t-t_{i,j})
\end{equation}
where $v_i(t) = [F_\gV(v_i)|| \mathrm{TDE}(t-t_i)]$ and $\alpha_k$ is attention weight computed as in GAT \citep{velivckovic2017graph}. $\mathrm{TDE}$ is a non-linear time difference embedding function such as Time2Vec \cite{kazemi2019time2vec}.
\cref{eq:merge} uses the embeddings of the nodes, edge time difference, and a non-linear $\mathrm{merge}$ function to give the final prediction.
\begin{equation}
\label{eq:merge}
    p_{i,j}(t) = \mathrm{merge}(z_i(t),z_j(t),z_{i,j}(t))
\end{equation}
\cref{eq:node_embedding,eq:edge_embedding,eq:merge}  constitute the prediction module of LDTGN. The prediction module only requires $t_i,t_j$ and $t_{i,j}$ from the memory module. Hence, these timestamps are the sole data of the memory module. In the experiments, we implemented LDTGN with a memory batch size of 1, thus eliminating the adverse effects associated with \textit{missing updates}. This design choice not only mitigates these negative impacts but also obviates the need for a message aggregator required in traditional TGNs. LDTGN operates with a minimal memory batch size and with a high throughput thanks to the removal of $\mathrm{msg}_s$,$\mathrm{msg}_d$ and $\mathrm{mem}$ from the memory module. Standard TGNs save states only for the nodes, but LDTGN also saves states for the edges. This does not add an additional memory to LDTGN over other TGNs since TGNs save the full graph to incorporate topological information in the prediction.

In the scenarios where the throughput is allowed to be smaller, and the \textit{missing updates} negative effects are neglectable for small memory batch size,
LDTGN can incorporate long-term dependencies. We refer to this variant of LDTGN as LDTGN-mem. To achieve long-term dependencies, LDTGN-mem is implemented with a heavier memory module. This memory module generates the following messages:
\begin{gather}
 m_i(t)= [s_i(t^-)||s_j(t^-)|| TDE(t-t_i)] \\
 m_j(t)= [s_j(t^-)||s_i(t^-)|| TDE(t-t_j)]  
\end{gather}
The aggregation function for the messages takes only the most recent message per node, and the $\mathrm{mem}$ function is set to be a $\mathrm{GRU}$ cell:
\begin{equation}
   s_i(t) = \mathrm{GRU}(\overline{m}_i(t),s_i(t^-))
\end{equation}
To incorporate the long-term memory in the prediction module, LDTGN-mem adds the current learned state to the data of each node:
\begin{equation}
    v_i(t) = [F_\gV(v_i)|| \mathrm{TDE}(t-t_i) || s_i(t^-)]
\end{equation}
In contrast to LDTGN, LDTGN-mem has to operate with a memory batch size larger than 1 to ensure a reasonable throughput. We chose to implement LDTGN-mem with a memory batch size of 50. This is because we observed earlier in \cref{fig:missing_updates} that the incidence of \textit{missing updates} with a batch size of 50 is not severe. In addition LDTGN-mem operates with an acceptable throughput when using this memory batch size. The adjustments required for the LDTGN-mem are detailed in the illustration of our model at \cref{fig:framework}.

\section{Experiments}
\label{sec:exp}

This section contains the description of the experiments we used to evaluate the performance of our model.
All the experiments were performed using DyGLib \citep{yu2023towards} -- the unified library for dynamic graph learning evaluation. DyGLib contains various real world datasets including large-scale dynamic graphs with millions of edges. The experiments are for future edge prediction with random negative edge sampling on the following datasets: Wikipedia, Reddit, MOOC, lastFM, Enron, Social Evo., UCI, Flights, Can. Parl., US Legis., UN Trade, UN Vote, and Contacts that were collected by \citet{poursafaei2022towards}. Additional information and statistics regarding the datasets can be found in \cref{appendix:A}.
We used seven well-known methods as baselines for the task of future edge prediction: DyRep \citep{trivedi2019dyrep}, TGAT \citep{xu2020inductive}, TGN \citep{rossi2020temporal}, CAWN \citep{wang2021inductive}, EdgeBank \citep{poursafaei2022towards}, GraphMixer \citep{cong2023we} and DyGFormer \citep{yu2023towards}. Additional
information regarding the baselines can be found in \cref{appendix:B}. 
We adopted the approach used in previous works and split the dataset into training, validation, and test sets by performing a chronological split of 70\%--15\%--15\%. We report the mean and standard deviation of the Average Precision (AP) on the test set. Results for Areas Under the
Receiver Operating Characteristic Curve (AUC-ROC) are detailed in \cref{appendix:D}.  
\subsection{Future edge prediction}
\label{sec:dynamic_link_prediction}
In the first experiment, we tested transductive future edge prediction with random negative edge sampling, i.e., for each positive edge in the datasets, a negative edge with the same source and a random destination is sampled. The results are presented in \cref{fig:tran}. We also performed an experiment for the inductive future edge prediction setting, in which all the edges in the validation and test sets must contain nodes that have not been previously seen in the training set. The results for this experiment are reported in \cref{fig:ind}.
The baselines' results were computed with DyGLib using the hyperparameters configurations as described in \citep{yu2023towards}. Additional implementation-specific details of LDTGN and LDTGN-mem and their training methodology are detailed in \cref{appendix:C}.
LDTGN achieves state-of-the-art or comparable results compared to the baselines for the setting of transductive and inductive future edge prediction.
In benchmarks where the negative effects of the \textit{missing updates} are insignificant for small batch sizes, LDTGN and LDTGN-mem achieve comparable performance to DyGFormer. In the benchmarks where \textit{missing updates} have substantial influence, such as US Legis, LDTGN considerably outperforms the compared baselines since it completely removes all the \textit{missing updates} when using a batch size of 1. 
\begin{table*}[htbp!]
\begin{center}
\caption{\label{fig:tran} AP for transductive future edge prediction with random negative
sampling over five runs. The significantly best result for each benchmark appears in bold font.}
\resizebox{\textwidth}{!}{
    \begin{tabular}{c c c c c c c c || c c}
     \toprule
     Dataset & DyRep & TGAT & TGN & CAWN & EdgeBank  & GraphMixer & DyGFormer & LDTGN (ours) & LDTGN-mem (ours)\\
     \midrule
    Wikipedia & 94.86±0.06  & 96.94±0.06 & 98.45±0.06 & 98.76±0.03 & 90.37±0.00 & 97.25±0.03 & \textbf{99.03±0.02} & 98.86±0.02 & \textbf{98.99±0.03}\\
    Reddit & 98.22±0.04 & 98.52±0.02 & 98.63±0.06 & 99.11±0.01 & 94.86±0.00 & 97.31±0.01 & 99.22±0.01 & 98.61±0.01 & \textbf{99.28±0.02}\\
    MOOC &81.97±0.49 & 85.84±0.15 & 89.15±1.60 & 80.15±0.25 & 57.97±0.00 & 82.78±0.15 & 87.52±0.49 & 83.34±1.47 & \textbf{91.73±0.65}\\
    lastFM &71.92±2.21  & 73.42±0.21 & 77.07±3.97 & 86.99±0.06 & 79.29±0.00 & 75.61±0.24 & \textbf{93.00±0.12} & 90.81±0.01 & 91.22±0.31\\
    Enron &82.38±3.36  & 71.12±0.97 & 86.53±1.11 & 89.56±0.09 & 83.53±0.00 & 82.25±0.16 & 92.47±0.12 & \textbf{98.10±0.01} & 92.28±0.32\\
    Social Evo. &88.87±0.30  & 93.16±0.17 & 93.57±0.17 & 84.96±0.09 & 74.95±0.00 & 93.37±0.07 & 94.73±0.01 & \textbf{95.45±0.51} & 94.02±0.16\\
    UCI &65.14±2.30 & 79.63±0.70 & 92.34±1.04 & 95.18±0.06 & 76.20±0.00 & 93.25±0.57 & 95.79±0.17 & \textbf{97.05±0.01} & 95.75±0.04\\
    Flights &95.29±0.72  & 94.03±0.18 & 97.95±0.14 & 98.51±0.01 & 89.35±0.00& 90.99±0.05 & \textbf{98.91±0.01} & 97.50±0.07 & 98.76±0.06\\
    Can. Parl. &66.54±2.76  & 70.73±0.72 & 70.88±2.34 & 69.82±2.34 &  64.55±0.00 & 77.04±0.46 & 97.36±0.45 & \textbf{99.47±0.03} & 72.82±9.17\\
    US Legis. &75.34±0.39  & 68.52±3.16 & 75.99±0.58 & 70.58±0.48 & 58.39±0.00 & 70.74±1.02 & 71.11±0.59 & \textbf{92.08±0.09} & 80.93±0.48\\
    UN Trade &63.21 ± 0.93  & 61.47±0.18 & 65.03±1.37 & 65.39±0.12 & 60.41±0.00 & 62.61±0.27 & 66.46±1.29 & \textbf{97.82±0.07} & 96.65±0.19\\
    UN Vote &62.81 ± 0.80  & 52.21±0.98 & 65.72±2.17 & 52.84±0.10 & 58.49±0.00 & 52.11±0.16 & 55.55±0.42 & \textbf{80.94±1.43} & 71.21±1.14\\
    Contacts &95.98 ± 0.15  & 96.28±0.09 & 96.89±0.56 & 90.26±0.28 &  92.58±0.00 & 91.92±0.03 & 98.29±0.01 & 98.19±0.03 & \textbf{98.78±0.04}\\
\bottomrule
\end{tabular}
    }
\end{center}
\end{table*}
\begin{table*}[htbp!]
\begin{center}
\caption{\label{fig:ind} AP for inductive future edge prediction with random negative
sampling over five different runs. The significantly best result for each benchmark appears in bold font.}
\resizebox{\textwidth}{!}{
    \begin{tabular}{c c c c c c c || c c}
     \toprule
     Dataset & DyRep & TGAT & TGN & CAWN & GraphMixer & DyGFormer & LDTGN (ours) & LDTGN-mem (ours)\\
     \midrule
Wikipedia &92.43±0.37  & 96.22±0.07 & 97.83±0.04 & 98.24±0.03 & 96.65±0.02 & 98.59±0.03 & \textbf{98.74±0.02} & 98.40±0.04\\
Reddit &96.09±0.11  & 97.09±0.04 & 97.50±0.07 & 98.62±0.01 & 95.26±0.02 & 98.84±0.02 & 98.00±0.04 & \textbf{98.86±0.02}\\
MOOC &81.07±0.44  & 85.50±0.19 & 89.04±1.17 & 81.42±0.24 & 81.41±0.21 & 86.96±0.43 & 82.73±1.52 & \textbf{90.61±0.32}\\
LastFM &83.02±1.48 & 78.63±0.31 & 81.45±4.29 & 89.42±0.07  & 82.11±0.42 & \textbf{94.23±0.09} & 92.17±0.01 & 92.62±0.59\\
Enron &74.55±3.95 & 67.05±1.51 & 77.94±1.02 & 86.35±0.51  & 75.88±0.48 & 89.76±0.34 & \textbf{96.06±0.09} & 88.07±0.56\\
Social Evo. &90.04±0.47 & 91.41±0.16 & 90.77±0.86 & 79.94±0.18  & 91.86±0.06 & 93.14±0.04 & \textbf{94.37±0.68} & 91.31±0.22\\
UCI &57.48±1.87 & 79.54±0.48 & 88.12±2.05 & 92.73±0.06  & 91.19±0.42 & 94.54±0.12 & \textbf{94.92±0.01} & 93.00±0.12\\
Flights &92.88±0.73& 88.73±0.33 & 95.03±0.60 & 97.06±0.02  & 83.03±0.05 & \textbf{97.79±0.02} & 95.60± 0.10 & 97.31±0.16\\
Can. Parl. &54.02±0.76 & 55.18±0.79 & 54.10±0.93 & 55.80±0.69  & 55.91±0.82 & 87.74±0.71 & \textbf{97.83±0.06} & 58.05±3.08\\
US Legis. &57.28±0.71 & 51.00±3.11 & 58.63±0.37 & 53.17±1.20  & 50.71±0.76 & 54.28±2.87 & \textbf{83.76±0.44} & 65.75±1.57\\
UN Trade &57.02±0.69& 61.03±0.18 & 58.31±3.15 & 65.24±0.21  & 62.17±0.31 & 64.55±0.62 & \textbf{97.43±0.07} & 89.21±1.15\\
UN Vote &54.62±2.22 & 52.24±1.46 & 58.85±2.51 & 49.94±0.45  & 50.68±0.44 & 55.93±0.39 & \textbf{81.29±1.41} & 63.54±2.09\\
Contacts &92.18±0.41 & 95.87±0.11 & 93.82±0.99 & 89.55±0.30  & 90.59±0.05 & \textbf{98.03±0.02} & 97.85±0.03 & \textbf{97.94±0.13}\\
\bottomrule
    \end{tabular}}
    \end{center}
    \end{table*}

\subsection{Memory and running time performance}
\begin{figure}[!htb]
   \begin{minipage}{0.49\textwidth}
     \centering
     \includegraphics[width=.99\linewidth]{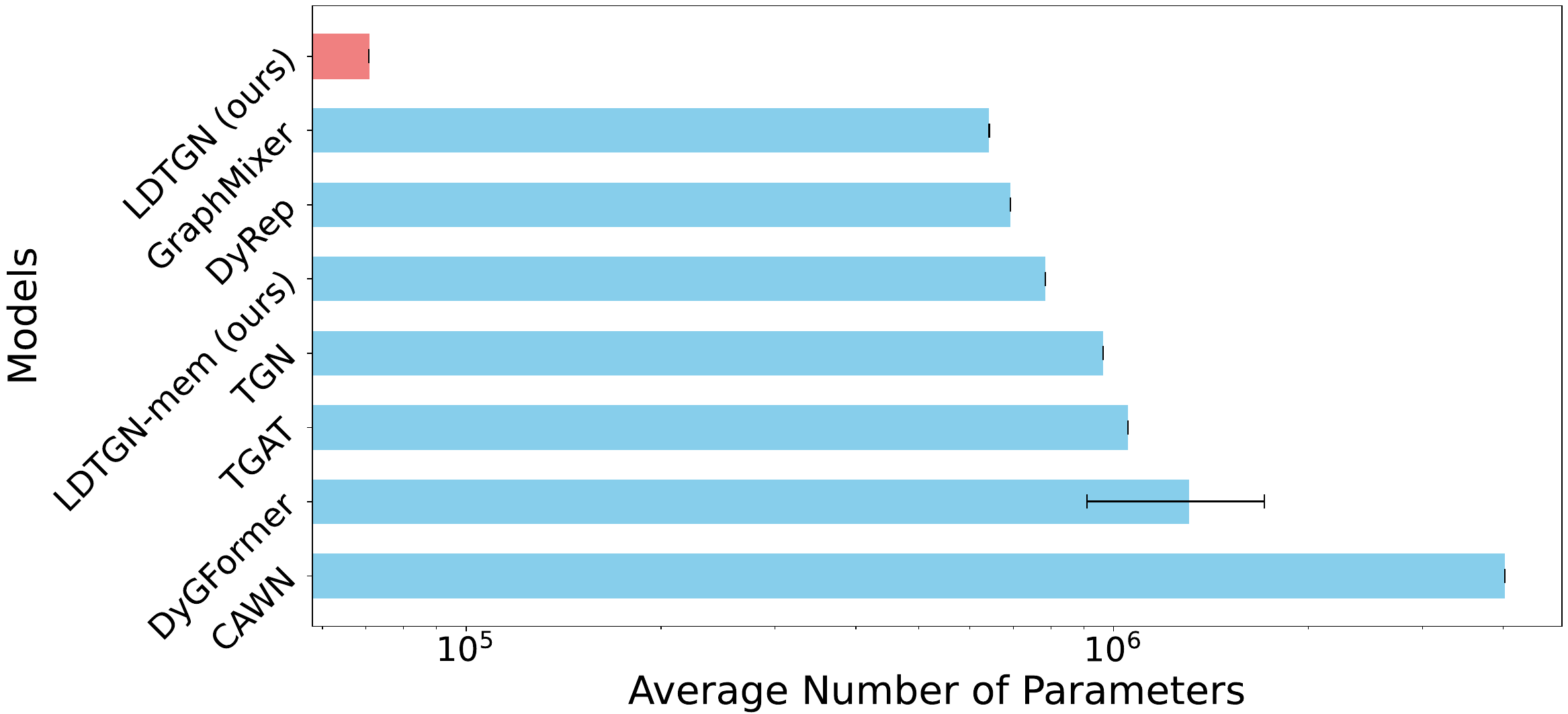}
     \caption{Average number of learnable parameters used by the baselines and our model. The black ranges indicate the standard deviation of the average number of learnable parameters.}
\label{fig:params}
   \end{minipage}\hfill
   \begin{minipage}{0.49\textwidth}
     \centering
     \includegraphics[width=.99\linewidth]{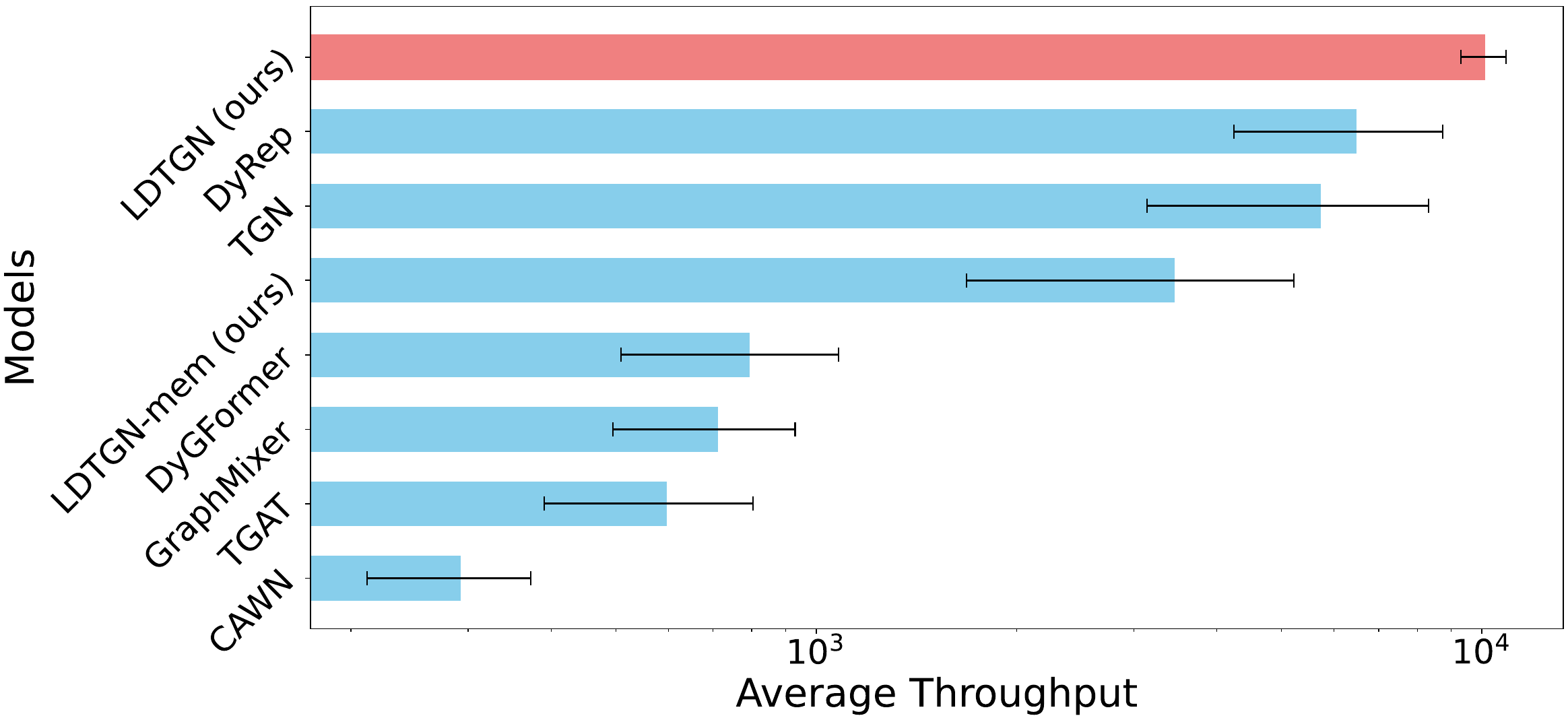}
     \caption{Average throughput (processed edges per second) of the baselines and our model. The black ranges indicate the standard deviation of the average throughput.}
\label{fig:throughput}
   \end{minipage}
\end{figure}

We calculated the average number of learnable parameters required for each model to achieve its best performance and reported it in \cref{fig:params}. We also measured the average throughput at inference time for each model over all the datasets, where the throughput is defined as the number of edges the model can process in a single second. The results are shown in \cref{fig:throughput}. In both \cref{fig:throughput} and \cref{fig:params}, LDTGN surpasses the other baselines by a large margin in terms of efficiency. Note that the throughput of 
 the baselines were measured using a batch size that is at least the batch size used for LDTGN; hence, the results in \cref{fig:throughput} are also proportionate to the latency of LDTGN compared to the other baselines.

\section{Conclusion}
In this work, we introduced the \textit{missing updates} phenomenon caused by using batches in memory-based models for dynamic graph learning. We showed a strict negative connection between the frequency of \textit{missing updates} in datasets and the performance of the memory-based models, causing a trade-off with respect to their batch size. To balance this trade-off, we presented the decoupling strategy for designing temporal graph networks. Decoupling enables two types of batches -- one for the memory module and the other for the prediction module. In this way, temporal graph networks can increase the frequency of the updates while still handling their arrival streams. In addition, we introduced LDTGN -- a lightweight model for the task of future edge prediction that is highly efficient in terms of time and memory. LDTGN can be equipped with a heavier memory module when possible, allowing it to better capture long-term dependencies. We also showed by extensive experiments that LDTGN has outstanding performance for both transductive and inductive tasks, achieving state-of-the-art or comparable performance on most of the tested benchmarks.
\clearpage
\bibliography{neurips_2024}
\bibliographystyle{neurips_2024}
\clearpage
\appendix
\section{Datasets statistics and descriptions}
\label{appendix:A}

In our experiments we used the following dynamic graph datasets:

• Wikipedia \citep{kumar2019predicting}: Wikipedia edit requests log over one month, where the editing users and Wikipedia pages
are represented as nodes and the edit requests are modeled as edges. The edges are timestamped and contain LIWC feature vectors \citep{pennebaker2001linguistic} of the requested text to post.

• Reddit \citep{kumar2019predicting}: Reddit post requests log over one month where the posting users and subreddits are represented as nodes and the posting requests are modeled as edges.

• MOOC \citep{kumar2019predicting}: Students' access records to MOOC online courses, where students and content units (e.g., videos, answers, etc.) are described as nodes and the access actions (viewing a video, submitting an answer, etc.) are modeled as edges. The edges are timestamped and have four features describing the action.

• LastFM \citep{kumar2019predicting}: LastFM listening records over one month, where the LastFM users and the songs are represented as nodes and there is an edge between the users and the songs to which they listened. The edges are timestamped and do not contain any features.

• Enron \citep{shetty2004enron}: Email logs of the Enron employees over a period of three years, where the employees are modeled as nodes and a single edge represents an email sent between two employees. The edges are timestamped and do not contain any features.

• Social Evo. \citep{madan2011sensing}: Documentation of the everyday life of undergraduate students living in dormitories from October 2008 to May 2009. Represented as a mobile phone proximity network where each edge has two features.

• UCI \cite{panzarasa2009patterns}: Messages logs of the online community of students from the University of California, Irvine, where the students are modeled as nodes and a single edge represents a message sent between two students. The edges are timestamped with a granularity of seconds.

• Flights \citep{strohmeier2021crowdsourced}: Tracked air traffic during the COVID-19 pandemic, where the airports are modeled as nodes and the edges are the tracked flights between two airports. The edges are timestamped and weighted. The weight of the edges indicates the number of flights between the airports in a day.

• Can. Parl. \citep{huang2020laplacian}: Documented interactions between
Canadian members of parliaments from 2006 to 2019, where the members of parliaments are described as nodes, two of which are connected by an edge if they both voted “yes” on a bill. The edges are timestamped and weighted. The weight of the edges indicates the number of times that one member voted “yes”  for another member's bill within one year.

• US Legis. \citep{fowler2006legislative}: Documented interactions in the US Senate, where legislators are modeled as nodes, where two of which are connected by an edge if they co-sponsored a bill. The edges are timestamped and weighted. The weight of the edges indicates the number of times that two members of the US Congress co-sponsored a bill in a given term.

• UN Trade \citep{macdonald2015rethinking}: Documented global food and agriculture trading connections spanning over 30 years, where nations are represented as nodes, two of which are connected by an edge if they have an agriculture import or export relations. The edges are timestamped and weighted. The weight of the edges is the sum of normalized
agriculture import or export values between two countries.

• UN Vote \citep{Voeten2009United}: Documentation of roll-call votes in the United Nations General Assembly from
1946 to 2020 where nations are represented as nodes, two of which are connected by an edge if they both voted “yes” for an item. The edges are timestamped and weighted. The weight of the edges is the number of times the two countries vote “yes” on a call.

• Contact \citep{sapiezynski2019interaction}: Physical proximity records documenting around
700 university students over a period of four weeks, where the students are modeled as nodes, two of which are connected by an edge if they each are within close proximity to each other. The edges are timestamped and weighted. The weight of the edges
specifies the physical proximity between two students.

The full statistics of the datasets as collected by \citet{yu2023towards} are reported in \cref{fig:statistics}.

\clearpage

\begin{table}[htbp]
\begin{center}
\caption{\label{fig:statistics}
    Datasets statistics.}
\resizebox{\textwidth}{!}{
    \begin{tabular}{c c c c c c c}
     \hline
     Dataset & Domain & \#Nodes & \#Edges & \#Edge Features & Bipartite & Duration\\
     \hline
    Wikipedia    & Social      & 9,227  & 157,474   & 172 & True  & 1 month       \\
    Reddit       & Social      & 10,984 & 672,447   & 172 & True  & 1 month       \\
    MOOC         & Interaction & 7,144  & 411,749   & 4   & True  & 17 months     \\
    LastFM       & Interaction & 1,980  & 1,293,103 & –   & True  & 1 month       \\
    Enron        & Social      & 184    & 125,235   & –   & False & 3 years       \\
    Social Evo.  & Proximity   & 74     & 2,099,519 & 2   & False & 8 months      \\
    UCI          & Social      & 1,899  & 59,835    & –   & False & 196 days      \\
    Flights      & Transport   & 13,169 & 1,927,145 & 1   & False & 4 months      \\
    Can. Parl.   & Politics    & 734    & 74,478    & 1   & False & 14 years      \\
    US Legis.    & Politics    & 225    & 60,396    & 1   & False & 12 terms \\
    UN Trade     & Economics   & 255    & 507,497   & 1   & False & 32 years      \\
    UN Vote      & Politics    & 201    & 1,035,742 & 1   & False & 72 years      \\
    Contact      & Proximity   & 692    & 2,426,279 & 1   & False & 1 month       \\
\hline
    \end{tabular}}
    \end{center}
    \end{table}

In \cref{tab:missing_updates1} we report the ratio of inputs that depend on at least a single \textit{missing update} in their 1-hop neighborhood. In \cref{tab:missing_updates2} we report the average number of \textit{missing updates} affecting the 1-hop neighborhood of the nodes. \cref{tab:missing_updates1,tab:missing_updates2} contain the \textit{missing updates} statistics for all the datasets used in this work for various batch sizes. 

\begin{table}[htbp!]
\centering
\begin{minipage}{0.48\textwidth}
\centering
\caption{\label{tab:missing_updates1}  Ratio of inputs that depend on at least a single \textit{missing update} in their 1-hop neighborhood.}
\resizebox{\linewidth}{!}{
    \begin{tabular}{c c c c c c c}
     \toprule
     Dataset & 1 & 10 & 25 & 50 & 100 & 200 \\
     \midrule
    Wikipedia & 0 &0.23 & 0.42 & 0.55 & 0.67 & 0.76 \\
    Reddit & 0 &0.31 & 0.52 & 0.67 & 0.78 & 0.86 \\
    MOOC & 0 &0.88 & 0.95 & 0.98 & 0.99 & 0.99 \\
    LastFM & 0 &0.74 & 0.88 & 0.94 & 0.97 & 0.98 \\
    Enron & 0 &0.85 & 0.92 & 0.95 & 0.98 & 0.99 \\
    Social Evo. & 0 &0.90 & 0.96 & 0.98 & 0.99 & 0.99 \\
    UCI & 0 &0.70 & 0.85 & 0.91 & 0.95 & 0.97 \\
    Flights & 0 &0.82 & 0.90 & 0.94 & 0.96 & 0.98 \\
    Can. Parl. & 0 &0.90 & 0.96 & 0.98 & 0.99 & 0.99 \\
    US Legis. & 0 &0.90 & 0.96 & 0.98 & 0.99 & 0.99 \\
    UN Trade & 0 &0.90 & 0.96 & 0.98 & 0.99 & 0.99 \\
    UN Vote & 0 &0.90 & 0.96 & 0.98 & 0.99 & 0.99 \\
    Contacts & 0 &0.86 & 0.94 & 0.97 & 0.98 & 0.99 \\
\bottomrule
    \end{tabular}}
\end{minipage}\hfill%
\begin{minipage}{0.48\textwidth}
\centering
\caption{\label{tab:missing_updates2} Average number of \textit{missing updates} affecting the 1-hop neighborhood of each input node.}
\resizebox{\linewidth}{!}{
    \begin{tabular}{c c c c c c c}
     \toprule
     Dataset & 1 & 10 & 25 & 50 & 100 & 200 \\
     \midrule
    Wikipedia & 0 & 0.25 & 0.65 & 1.19 & 2.02 & 3.28 \\
    Reddit & 0 & 0.15 & 0.39 & 0.78 & 1.53 & 2.99 \\
    MOOC & 0 & 0.72 & 1.45 & 2.38 & 3.90 & 6.57 \\
    LastFM & 0 & 0.45 & 1.16 & 2.10 & 3.59 & 5.93 \\
    Enron & 0 & 2.95 & 6.07 & 9.81 & 15.55 & 24.95 \\
    Social Evo. & 0 & 1.86 & 3.15 & 5.22 & 9.77 & 19.22 \\
    UCI & 0 & 0.95 & 2.12 & 3.67 & 5.98 & 9.31 \\
    Flights & 0 & 2.82 & 5.77 & 8.79 & 11.97 & 14.55 \\
    Can. Parl. & 0 & 4.41 & 11.46 & 22.33 & 40.65 & 66.87 \\
    US Legis. & 0 & 4.19 & 10.10 & 17.41 & 25.26 & 31.09 \\
    UN Trade & 0 & 4.37 & 11.21 & 21.35 & 37.41 & 57.00 \\
    UN Vote & 0 & 3.75 & 8.56 & 14.35 & 21.47 & 28.01 \\
Contacts & 0 & 1.74 & 2.60 & 3.15 & 3.85 & 5.13 \\
\bottomrule
    \end{tabular}}

\end{minipage}
\end{table}

\clearpage

\section{Baselines descriptions}
\label{appendix:B}

We used the following baselines for the evaluation experiments of dynamic graph learning:

•DyRep \citep{trivedi2019dyrep}: DyRep is an RNN-based architecture that utilizes a temporal attention mechanism to exploit the dynamic structure of the graphs.

• TGAT \citep{xu2020inductive}: TGAT uses a time-encoding function and aggregates neighborhood information using self-attention to compute the embedding for each node.

• TGN \citep{rossi2020temporal}: TGN is a general architecture for CTDG learning tasks. It uses both a prediction module and a memory module to get relevant and accurate predictions for each input at each moment in time. It does this by aggregating information from the neighborhood of each node and maintain learnable updated memory which is based on RNN, and thus also solves the staleness problem.

• CAWN \citep{wang2021inductive}: The CAWN model is based on causal anonymous walks that are generated for each node. The walks are encoded using RNNs and aggregated to achieve the node representation.

• EdgeBank \citep{poursafaei2022towards}:  EdgeBank is a memorization algorithm that saves any seen update and, given an input, it predicts
according to a simple decision rule that can be one of the following: whether the input was seen in
the last few iterations (EdgeBank$_{th}$) or in the last few time units (EdgeBank$_{tw}$), or whether the input has already been seen a sufficient number of times (EdgeBank$_{re}$). While EdgeBank can also have a decision rule that is based on infinite memory i.e., predicts positive for any previously seen edge and predicts negative otherwise (EdgeBank$_{\inf}$).
The algorithm's simplicity allows it to perform extremely fast, making it significantly faster than any other model for dynamic graph learning. In our experiments, we report the best results of EdgeBank among all of its decision rule variations.

• GraphMixer \citep{cong2023we}: GraphMixer uses three components for the task of future edge prediction: a link-encoder that is 
based on MLP and fixed time-encoding function, a node-encoder that  only performs neighborhood mean-pooling and another MLP for edge prediction.

• DyGFormer \citep{yu2023towards}: DyGFormer is a transformer-based architecture. To generate an encoding for a given interaction, DyGFormer generates a co-occurrence embedding of the interaction in addition to a neighborhood representation for each interacting node.  Then it uses a patching technique on historical representations of the interacting nodes to better capture long-term temporal dependencies. The patches are then sent to a transformer and its outputs are averaged to create the final representation.

\section{Additional Implementation Details}
\label{appendix:C}

\subsection{Supporting additional update types}

In \cref{subsec:ldtgn} we described how to handle an edge addition update. To further support the update of the removal of the edge $e_{i,j}$, $t_{i,j}$ should be set to $0$. Similarly, when a node addition update of the node $v_i$ occurs, $t_i$ should be set to the current time. $t_i$ should be set to $0$ when it is removed.

\subsection{Node classification}
 To adjust LDTGN for dynamic node classification, the $\mathrm{MERGE}$ function needs to be removed, s.t., the prediction operation is applied directly on the node embedding. The Wikipedia dataset can also be used for dynamic node classification, therefore we used it to evaluate our model compared to other baselines: 

 \begin{table*}[htbp!]
\begin{center}
\caption{\label{fig:node}  AUC-ROC for node prediction task on the Wikipedia dataset.}
\resizebox{\textwidth}{!}{
    \begin{tabular}{c c c c c c c c c c}
     \toprule
     Dataset & DyRep &	TGAT &	TGN&	CAWN&	GraphMixer&	DyGFormer&	LDTGN (ours)\\
     \midrule
    Wikipedia & 86.39±0.98&	84.09±1.27&	86.38±2.34&	84.88±1.33&	86.80±0.79&	87.44±1.08&	86.71±0.44\\
\bottomrule
\end{tabular}
    }
\end{center}
\end{table*}
For this task of node classification LDTGN achieves comparable performance to previous state-of-the-art while still being the most efficient in terms of throughput and latency.

\subsection{Further implementation details}

For the $\mathrm{TDE}$ of LDTGN we used an $\mathrm{MLP}$ with two hidden layers and two activation layers of ReLU \citep{nair2010rectified}. Each linear layer of $\mathrm{TDE}$  outputs vector of length 100. Before applying $\mathrm{TDE}$ the time difference need to be normalised to ease the learning process. We used \cref{eq:norm} to normalise the time difference, where $C$ is the length of the dataset.

\begin{equation}
\label{eq:norm}
    \mathrm{normalise}(t) = \frac{log(1+t)}{log(1+C)}
\end{equation}

For LDTGN-mem, we used Time2Vec as the $\mathrm{TDE}$ function. Time2Vec utilizes the cosine function, thus omitting the need for normalization.

The $\textrm{merge}$ function of LDTGN and LDTGN-mem is an $\mathrm{MLP}$ that maps multiple input vectors into a single value that represents the probability of the edge to be positive. The $\mathrm{MLP}$ first applies linear layer that maps the three vectors to a single vector. Then reduces the vector's dimension to 80, 10 and finally to 1. After each dimensionality reduction, a ReLU is being applied. Finally, a sigmoid function is applied on the result to obtain the probability of an edge to be positive.

In practice, it is challenging to utilize the full neighborhood of input nodes to compute the predictions and withstand a reasonable throughput, since the neighborhood of each node is expected to grow overtime. Thus, we implemented our models using the recent neighbors sampling strategy that was suggested by \citet{rossi2020temporal} in which only the $k$ neighbors of each hop which were recently involved in an update are used for computing the predictions. For our models, we used $k=20$.

\subsection{Training}
We trained the models for 100 epochs with a patience of 20 epochs before early stopping. We used binary cross entropy loss as the objective function and optimized the models using Adam's algorithm with a learning rate of $10^{-4}$. \\ \\All the experiments were performed on Intel(R) Xeon(R) Gold 6130 CPU @ 2.10GHz and NVIDIA GeForce RTX 3090.

\section{Additional results}
\label{appendix:D}

In \cref{fig:tran_auc} and \cref{fig:ind_auc} we report the AUC-ROC of our proposed model and baselines for the transductive and inductive future edge prediction tasks, respectively.

\begin{table*}[htbp]
\begin{center}
\caption{\label{fig:tran_auc} : AUC-ROC for transductive future edge prediction with random negative
sampling over five runs. The significantly best result for each benchmark appears in bold font.}
\resizebox{\textwidth}{!}{
    \begin{tabular}{c c c c c c c c || c c}
     \toprule
     Dataset & DyRep  & TGAT & TGN & CAWN & EdgeBank & GraphMixer & DyGFormer & LDTGN (ours) & LDTGN-mem (ours)\\
     \midrule
Wikipedia  & 94.37 ± 0.09  & 96.67 ± 0.07  & 98.37 ± 0.07  & 98.54 ± 0.04  & 90.78 ± 0.00  & 96.92 ± 0.03  & \textbf{98.91 ± 0.02} & 98.67±0.01 & \textbf{98.90±0.05}\\ 
Reddit & 98.17 ± 0.05  & 98.47 ± 0.02  & 98.60 ± 0.06  & 99.01 ± 0.01  & 95.37 ± 0.00  & 97.17 ± 0.02  & 99.15 ± 0.01 & 98.20±0.02 & \textbf{99.25±0.02}\\ 
MOOC & 85.03 ± 0.58  & 87.11 ± 0.19  & 91.21 ± 1.15  & 80.38 ± 0.26  & 60.86 ± 0.00  & 84.01 ± 0.17  & 87.91 ± 0.58 & 82.43±1.72 & \textbf{93.33±0.54}\\ 
LastFM &71.16 ± 1.89  & 71.59 ± 0.18  & 78.47 ± 2.94  & 85.92 ± 0.10  & 83.77 ± 0.00  & 73.53 ± 0.12  & \textbf{93.05 ± 0.10} & 90.79±0.01 & 91.68±0.51\\ 
Enron & 84.89 ± 3.00  & 68.89 ± 1.10  & 88.32 ± 0.99  & 90.45 ± 0.14  & 87.05 ± 0.00  & 84.38 ± 0.21  & 93.33 ± 0.13 & \textbf{98.31±0.01} & 93.35±0.42\\ 
Social Evo. & 90.76 ± 0.21  & 94.76 ± 0.16  & 95.39 ± 0.17  & 87.34 ± 0.08  & 81.60 ± 0.00  & 95.23 ± 0.07  & 96.30 ± 0.01 & \textbf{96.82±0.25} & 95.93±0.06\\ 
UCI & 68.77 ± 2.34  & 78.53 ± 0.74  & 92.03 ± 1.13  & 93.87 ± 0.08  & 77.30 ± 0.00  & 91.81 ± 0.67  & 94.49 ± 0.26 & \textbf{96.22±0.03} & 94.79±0.07\\ 
Flights & 95.95 ± 0.62  & 94.13 ± 0.17  & 98.22 ± 0.13  & 98.45 ± 0.01  & 90.23 ± 0.00  & 91.13 ± 0.01  & \textbf{98.93 ± 0.01} & 96.98±0.09 & 98.82±0.07\\ 
Can. Parl. &73.35 ± 3.67 & 75.69 ± 0.78  & 76.99 ± 1.80  & 75.70 ± 3.27  & 64.14 ± 0.00  & 83.17 ± 0.53  & 97.76 ± 0.41 & \textbf{99.68±0.02} & 77.66±7.92\\ 
US Legis. & 82.28 ± 0.32  & 75.84 ± 1.99  & 83.34 ± 0.43  & 77.16 ± 0.39  & 62.57 ± 0.00  & 76.96 ± 0.79  & 77.90 ± 0.58 & \textbf{94.88±0.10} & 87.96±0.53\\ 
UN Trade & 67.44 ± 0.83  & 64.01 ± 0.12  & 69.10 ± 1.67  & 68.54 ± 0.18  & 66.75 ± 0.00  & 65.52 ± 0.51  & 70.20 ± 1.44 & \textbf{97.91±0.06} & 97.16±0.17\\ 
UN Vote &67.18 ± 1.04  & 52.83 ± 1.12  & 69.71 ± 2.65  & 53.09 ± 0.22  & 62.97 ± 0.00  & 52.46 ± 0.27  & 57.12 ± 0.62 & \textbf{86.81±0.87} & 77.33±1.04\\ 
Contact & 96.48 ± 0.14  & 96.95 ± 0.08  & 97.54 ± 0.35  & 89.99 ± 0.34  & 94.34 ± 0.00  & 93.94 ± 0.02  & 98.53 ± 0.01 & 98.58±0.01 & \textbf{99.06±0.04}\\ 
\bottomrule
\end{tabular}
    }
    \end{center}
    \end{table*}

\begin{table*}[htbp]
\begin{center}
\caption{\label{fig:ind_auc} : AUC-ROC for inductive future edge prediction with random negative
sampling over 5 different runs. The significantly best result for each benchmark appears in bold font.}
\resizebox{\textwidth}{!}{
    \begin{tabular}{c c c c c c c || c c}
     \toprule
     Dataset & DyRep & TGAT & TGN & CAWN & GraphMixer & DyGFormer & LDTGN (ours) & LDTGN-mem (ours)\\
     \midrule
Wikipedia &91.49±0.45  & 95.90±0.09  & 97.72±0.03  & 98.03±0.04  & 95.57±0.20  & \textbf{98.48±0.03} & 98.23±0.00 & 98.30±0.06\\ 
Reddit & 96.05±0.12  & 96.98±0.04  & 97.39±0.07  & 98.42±0.02  & 93.80±0.07  & \textbf{98.71±0.01} & 97.30±0.03 & 98.56±0.05\\ 
MOOC &84.03±0.49 & 86.84±0.17  & 91.24±0.99  & 81.86±0.25  & 81.43±0.19  & 87.62±0.51 & 81.88±1.74 & \textbf{92.36±0.30}\\ 
LastFM &82.24±1.51  & 76.99±0.29  & 82.61±3.15  & 87.82±0.12  & 70.84±0.85  & \textbf{94.08±0.08} & 91.75±0.01 & 92.57±0.86\\ 
Enron &76.34±4.20 & 64.63±1.74  & 78.83±1.11  & 87.02±0.50  & 72.33±0.99  & 90.69±0.26 & \textbf{95.77±0.13} & 88.46±0.79\\ 
Social Evo. &91.18±0.49  & 93.41±0.19  & 93.43±0.59  & 84.73±0.27  & 93.71±0.18  & 95.29±0.03 & \textbf{96.03±0.37} & 94.01±0.2\\ 
UCI & 58.08±1.81  & 77.64±0.38  & 86.68±2.29  & 90.40±0.11  & 84.49±1.82  & 92.63±0.13 & \textbf{92.83±0.02} & 90.83±0.21\\ 
Flights & 93.56±0.70  & 88.64±0.35  & 95.92±0.43  & 96.86±0.02  & 82.48±0.01  & \textbf{97.80±0.02} & 94.44±0.21 & 97.39±0.21\\ 
Can. Parl. &55.27±0.49  & 56.51±0.75  & 55.86±0.75  & 58.83±1.13  & 55.83±1.07  & 89.33±0.48 & \textbf{98.73±0.05} & 58.59±4.42\\ 
US Legis. & 61.07±0.56  & 48.27±3.50  & 62.38±0.48  & 51.49±1.13  & 50.43±1.48  & 53.21±3.04 & \textbf{88.19±0.24} & 72.45±1.31\\ 
UN Trade &58.82±0.98  & 62.72±0.12  & 59.99±3.50  & 67.05±0.21  & 63.76±0.07  & 67.25±1.05 & \textbf{97.47±0.07} & 90.26±1.28\\ 
UN Vote &55.13±3.46  & 51.83±1.35  & 61.23±2.71  & 48.34±0.76  & 50.51±1.05  & 56.73±0.69 & \textbf{86.99±0.86} & 68.99±1.66\\ 
Contact &91.89±0.38  & 96.53±0.10  & 94.84±0.75  & 89.07±0.34  & 93.05±0.09  & \textbf{98.30 ± 0.02} & 98.26±0.02 & 98.22±0.14\\ 
\bottomrule
    \end{tabular}}
    \end{center}
    \end{table*}
\section{Decoupling potential speedup analysis}
\label{appendix:E}
In this section we analyze the potential speedup that models can achieve by using the decoupling strategy.   
The decoupling strategy does not affect running time directly, but rather aids to accelerate the running time of models without compromising their precision. \cref{fig:decoupled_running_time} demonstrates this exact idea: one can decouple a temporal model and increase its batch size significantly while maintaining a constant memory batch size. This will lead to running time improvement without compromising the precision of the decoupled model. In the context of a sequence containing updates for a model to apply and inputs for it to predict, denote the time it takes for the memory module to apply all the given updates in the sequence as $t_{memory}$ and denote the time it takes for the prediction module to finish computing the predictions for all the inputs in the sequence as $t_{prediction}$. The total time it takes for the model to finish to process the sequence is: 
\begin{equation}
    T_{total}=t_{memory}+t_{prediction}
\end{equation}
By decoupling the network one can reduce this time to
\begin{equation}
    T_{decouple\_total} = \frac{t_{prediction}}{BS_{new}/BS_{old}} + t_{memory}
\end{equation}
Where $BS_{old}$ is the batch size of the model before decoupling and $BS_{new}$ is the batch size of the decouple model.
Hence the potential speedup of the model is:
\begin{equation}
  speedup = \frac{BS_{new}\cdot t_{prediction} +BS_{new}\cdot t_{memory}}{BS_{old}\cdot t_{prediction} + BS_{new} \cdot t_{memory}}
\end{equation}

\end{document}